\documentclass[11pt]{article}
\usepackage[margin=1in]{geometry}
\usepackage{lmodern}             
\usepackage[T1]{fontenc}
\usepackage[utf8]{inputenc}
\usepackage{amsmath, amssymb}    
\usepackage{graphicx}            
\usepackage{booktabs, multirow}  
\usepackage{enumitem}            
\usepackage{caption, subcaption} 
\usepackage{natbib}              
\usepackage{hyperref}            
\usepackage{xcolor}              
\usepackage{changepage}          
\usepackage{algorithm}
\usepackage{algpseudocode}
\usepackage{amsmath}
\DeclareMathOperator*{\argmax}{argmax}
\usepackage{float}
\makeatletter
  \newlength{\OLDfptop}
\makeatother

\definecolor{amethyst}{rgb}{0.6, 0.4, 0.8}

\title{\bfseries Unsupervised Machine Learning for Scientific Discovery: Workflow and Best Practices}

\author{
  Andersen Chang\textsuperscript{$1$}\thanks{These authors contributed equally.} ,
  Tiffany M.\ Tang\textsuperscript{$2$}\footnotemark[1] ,
  Tarek M.\ Zikry\textsuperscript{$3,4$}\footnotemark[1] ,
  Genevera I.\ Allen\textsuperscript{$3,4$\thanks{Corresponding author: \texttt{genevera.allen@columbia.edu}}} \\[1em]
  \small\textsuperscript{1}Department of Neuroscience, Baylor College of Medicine\\
  \small\textsuperscript{2}Department of Applied and Computational Mathematics and Statistics, University of Notre Dame\\
  \small\textsuperscript{3}Department of Statistics, Columbia University\\
  \small\textsuperscript{4}Zuckerman Institute \& Irving Institute, Columbia University
}
\date{}

\begin{document}
\maketitle

\begin{abstract}
Unsupervised machine learning is widely used to mine large, unlabeled datasets to make data-driven discoveries in critical domains such as climate science, biomedicine, astronomy, chemistry, and more. However, despite its widespread utilization, there is a lack of standardization in unsupervised learning workflows for making reliable and reproducible scientific discoveries. In this paper, we present a structured workflow for using unsupervised learning techniques in science. We highlight and discuss best practices starting with formulating validatable scientific questions, conducting robust data preparation and exploration, using a range of modeling techniques, performing rigorous validation by evaluating the stability and generalizability of unsupervised learning conclusions, and promoting effective communication and documentation of results to ensure reproducible scientific discoveries. To illustrate our proposed workflow, we present a case study from astronomy, seeking to refine globular clusters of Milky Way stars based upon their chemical composition. Our case study highlights the importance of validation and illustrates how the benefits of a carefully‐designed workflow for unsupervised learning can advance scientific discovery.
\end{abstract}

\vspace{1em}

reproducibility, astronomy



\maketitle

\section{Introduction} \label{sec:intro}

Machine learning has emerged as an important tool in deriving data-driven insights, enabling researchers to extract meaningful patterns from large and complex data that would have been difficult for traditional methods to identify. As scientific problems increasingly involve high-dimensional datasets with nonlinear relationships and multivariate interactions, machine learning has become a critical tool for modeling complex relationships, identifying hidden structures, and making predictions in data science problems \citep{Roscher2020, Molnar2020}. Beyond improving accuracy and efficiency, machine learning enables novel forms of scientific discovery, such as identifying unexpected correlations, generating hypotheses, and simulating outcomes under diverse scenarios in scientific research contexts across multiple disciplines, including biology, chemistry, economics, and artificial intelligence. Due to this pervasion, efforts have been made amongst those in the data science community to establish general, model-agnostic frameworks for machine learning; this is important to ensure that novel findings are reproducible and reliable \citep{allen2023interpretable, samuel2020machine, simmhan2009building} and for improving consistency and effectiveness of collaborative work \citep{zahid2018enhancing, biswas2022art}, especially with non-statisticians.

In statistical machine learning, the approaches used for modeling in data analysis problems can be broadly categorized into two different classes. The first of these, known as \textit{supervised} learning, is defined by the existence of an explicitly-defined output or response variable in the data set. Machine learning techniques for supervised learning use the response variable as a quantitative target of modeling, finding mappings of the input or predictor variables that minimizes the error between the expected value of the response based on the observed predictor variables and the actual realized values of the response variable. The second class of approaches, known as \textit{unsupervised} learning, attempt to make discovery without a pre-defined response variable. Instead, these methods find hidden relationships and patterns in data by constructing representations of the structure of observations and features that allow for the discovery of correlations, groupings, and latent spaces in high-dimensional settings. Our goal in this paper is to provide collaborative data scientists and biostatisticians with a framework for creating workflows for the latter of these, i.e. unsupervised learning methods, with a particular focus on workflows for projects for generating data-driven discoveries using these methods.

\subsection{Unsupervised Learning for Discovery}

Unsupervised learning methods are a critical component of any data science or machine learning pipeline. In practice, unsupervised learning techniques are typically applied in supervised learning workflows for the purposes of data exploration, data visualization, feature engineering, preprocessing, and data preparation. However, unsupervised learning can also be a crucial tool for generating novel data-driven discoveries from unlabeled data.  Examples of major unsupervised tasks include clustering to discover group structures \citep{Xu2008}, graphical models to discover relationships \citep{Hastie2009}, anomaly detection to discover informative outliers \citep{Chandola2009}, and dimension reduction to discover low-dimensional latent subspaces capturing major patterns in high-dimensional data \citep{VanDerMaaten2009}. These unsupervised learning methods have been applied for generating novel, important data-driven discoveries in many different types of data science research problems in a diverse range of fields. For instance, clustering methods have been applied in genetics studies for identifying common gene expression patterns \citep{do2008clustering, hong2020rna}, in astrostatistics to investigate the distribution of mass and galaxies in the universe \citep{Fraix-Burnet_Bouveyron_Moultaka, Hawkins_Jofré_Masseron_Gilmore_2015, janvrin2014making}. Network analyses have utilized graphical models to study the structure of pathways underlying the connectome \citep{bullmore2011brain, fornito2013graph, gastner2016topology} and the organization of functional circuits \citep{zhu2018decoding, farahani2019application, weis2022unsupervised} within the brain, to find interactions in RNA sequences and interactions in gene expression in disease pathologies \citep{sinoquet2014probabilistic, lauritzen2003graphical}, and to study the movement of air particulate matter \citep{ebertuphoff2012causal, hu2016analysis}. Also, latent space estimates derived from applying dimension reduction methods have revealed spatiotemporal changes in climate patterns \citep{gamez2004nonlinear, sarhadi2017advances, tibau2021spatio}, to estimate novel drug-disease interactions \citep{purkayastha2019drug, wang2020novel, kim2023predicting, chow2022predicting}, and to uncover common biochemical signatures in gut microbiome samples \citep{li2022machine, armstrong2022applications}. These examples show that unsupervised learning can be an extremely useful tool in generating data-driven discoveries, and thus should be a foundational part of the literature in data science practice.

\subsection{Workflows}
 Though unsupervised discovery is an extremely important facet of modern data science and machine learning, there is currently a lack of well-established best practices and specific workflow procedures for producing robust, reproducible discoveries using unsupervised learning. Several works in the literature have studied and proposed workflows and practices for specific types of unsupervised learning tasks, including for clustering \citep{greene2008unsupervised}, autoencoder neural networks \citep{dike2018unsupervised, becker1991unsupervised}, natural language processing \citep{solan2005unsupervised}, dimension reduction \citep{huang2022towards, andersen2024supervised}, and image processing \citep{olaode2014unsupervised, raza2021tour}, as well as for specific fields such as single-cell multiomics \citep{chua2023workflow, bravo2019omics} and pathology \citep{roohi2020unsupervised, silburt2022morphious}. However, there is currently a dearth of existing resources concerning data science analytical pipelines for unsupervised discovery that are broadly applicable across variable techniques and domains. The lack of well-established general unsupervised discovery workflows is due in large part to the substantial number of unique challenges that arise from unsupervised learning for scientific discovery. One of these primary challenges is that, by nature, unsupervised learning problems do not include a target outcome that can be used as an objective for quantifying the accuracy of model fits, thereby complicating the tasks of model comparison and hyperparameter selection. For many different classes of unsupervised learning problems, there also does not exist the concept of a null distribution for which quantitative results can be compared against, thus complicating the task of post-hoc inference and model selection \citep{pick2023describing, watson2023philosophy}. Additionally, several unsupervised learning methods do not provide a mechanism for mapping new data points to a pre-existing model fit, which precludes the ability to use a training-test split for external validation of model performance \citep{sainburg2021parametric, shi2005robust}. Furthermore, recent work has shown that interpretations from widely used unsupervised learning methods within clustering and dimension reduction are largely unreliable and yield unstable results, underscoring a greater need for emphasis on validation and robustness within workflows \citep{gan2025machine}.

In this paper, we address the gap in guidance on how to create unsupervised learning workflows for scientific discovery. We first present a generalized, model-agnostic data science workflow for unsupervised learning, with specific recommendations and considerations for the challenges of unsupervised scientific discovery, and also provide a concise discussion of best practices in unsupervised analysis.  We then illustrate our workflow recommendations through an extensive case study from astronomy on discovering common origins of stars based on their chemical signatures and show how decisions made about the procedures in the workflow can affect resulting scientific discoveries.  


\section{Workflow for Unsupervised Discovery}\label{sec:workflow}
\begin{figure}[!ht]
    \centering
    \includegraphics[width=1\linewidth]{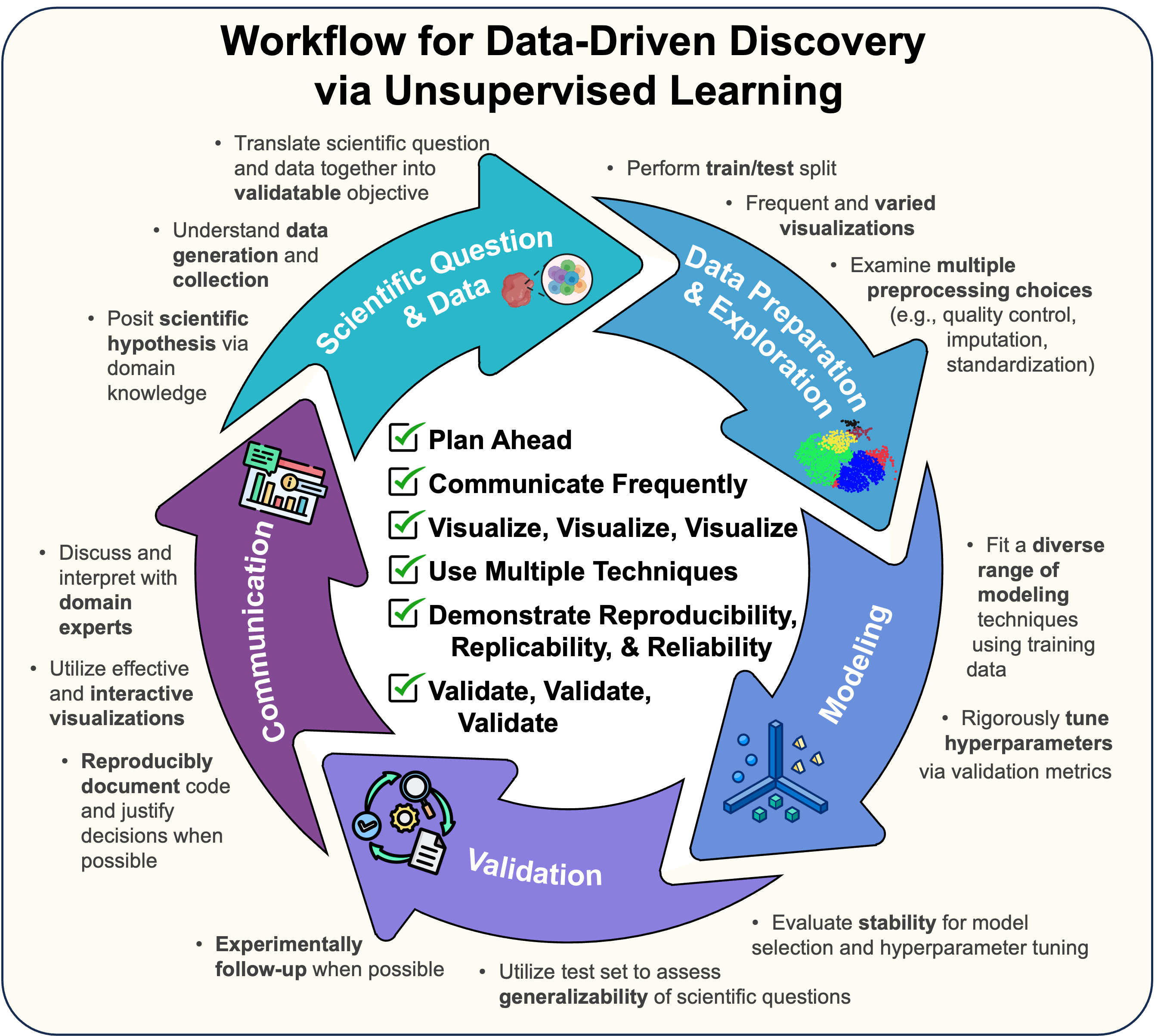}
    \caption{Illustration of Best Practices for Data-Driven Discovery via Unsupervised Learning. We summarize best practices, starting with generating validatable scientific questions, through robust data preparation and exploration, using a range of modeling techniques, evaluating the stability and generalizability of unsupervised learning conclusions, and the effective communication and documentation of results.}
    \label{fig:schema}
\end{figure}

In this section, we present a model-agnostic workflow for unsupervised learning for the task of producing data-driven discoveries. We note that many of the aspects of the workflow we discuss below will follow that of a standard data science pipeline \citep{koivisto2019efficient, munappy2020data, tiwari2007workflow, ma2016omics,Yu2020}. However, we will provide more detailed discussion into aspects that are specific to working with unsupervised learning methods. A summary of our suggested best practices can be seen in Figure~\ref{fig:schema}.

\subsection{Workflow} \label{sub:work}

\subsubsection{Scientific Question \& Data} 

The first step in creating an unsupervised learning workflow is to formulate the scientific question to be answered and to determine the type of data that will be collected or used to answer this question. It is important here to begin with a thorough literature review and to discuss with domain experts. This will help provide an understanding of the previous research in the field of study, ensure that the project addresses an existing gap in knowledge guidance in the field, and help position potential findings within the overall current knowledge in the literature \citep{frank2014doing}. 
Moreover, the scientific question should be formulated in tandem with a detailed understanding of the data generation or collection procedures.
Though problem formulation and data collection are typically considered two different aspects of a data science workflow, we emphasize that both of these facets need to be determined concurrently in order to ensure that the two are congruent for meeting the overall goals of the research project. 

For data collection, it is essential that the number of observations and the features that will be needed for the entirety of the pipeline be determined before any work is done. If new data will be gathered for the workflow, enough observations should be collected such that there is adequate statistical power for all future modeling and validation procedures. If instead the workflow will be utilizing data previously collected, then similar datasets from different sources can potentially be used to augment the available observations or applied as a test set for validation. Also, for both cases, the features to be measured should be thoroughly planned as well in order to ensure that the data are appropriate for answering the scientific questions; for example, potential confounding variables should be considered in order to aid with interpretations of findings \citep{jager2008confounding}.

Specification of the scientific question is pivotal for defining the overall direction of the workflow and research project as a whole. Furthermore, in unsupervised settings, the key in this step is to translate high-level research objectives into scientific questions that are \textbf{vaildatable}, i.e. with actionable objectives and quantifiable evaluation metrics, and which are answerable with the data that are available for the current project. For example, a broad research goal could be to explore the biological characteristics associated with observed incidences and progression of a particular disease using multiomics data, physiological testing results, and clinical notes. This overall project objective could be accordingly parsed and converted into several specific research questions involving unsupervised discovery, such as:

\begin{itemize}
    \item Are there differing phenotypic subgroups amongst the full set of disease cases? How many of these subgroups can we reliably find? How do the subgroups differ in terms of physiology or responses to treatment?
    \item Which features significantly covary amongst the cases? Are there physiological features that can be used as an indicator of genetic susceptibility?
\end{itemize}  
Notably, these questions are addressable by the data modalities that are available, and have clear quantifiable or objective measures for modeling results. On the other hand, a question such as \textit{"Are there covarying features associated with vulnerability?"} would not be a particularly useful scientific question, as "vulnerable subject" in this context needs to be re-defined in such a way that there is a quantifiable metric, e.g. time to death or level of symptoms.  Finally, we note that developing a validatable scientific question is a critical component, but can be especially challenging for unsupervised learning, as there are fewer objective measures for assessing results.

Before delving into discussions of the further parts of the unsupervised learning workflow, we pause to stress that the entire plan for modeling and validation should be planned out in this step as well. Though unforeseen circumstances may require adjustments to the workflow as steps are performed, creating an overall initial plan before starting the analysis provides several important advantages. First, this will help prevent critical incidents, such as failing to collect important features or not creating a train-test split in advance, that may require redoing parts of the workflow or collecting additional data. Secondly, this will help to ensure that all parts of the workflow are coherent with one another and with the overall research goals, e.g. that the results produced by the modeling steps actually address the scientific question, or that engineered features are compatible for use in modeling and validation procedures. Lastly, setting a plan will allow for all stakeholders in the project to know what to expect from the data analysis and to provide feedback before work begins.

\subsubsection{Data Preparation and Exploration} 

Once data have been collected, data preparation and exploration should be performed before modeling in order to ensure computational feasibility, to improve overall modeling results, and to aid in understanding and contextualization of findings. While we have separated these two topics into separate subsections, preparation and exploration should be a cyclical, closed-loop process that occurs across multiple iterations, as the latter will inform the user on the potential extra that needs to be done in the former.

\paragraph{Preparation.}

Typically, preprocessing of the raw data will need to be conducted in order to prepare the data for modeling. Firstly, the data should be randomly split into training and test sets for validation purposes before any other step. Next, data cleaning procedures should be applied; common tasks here include fixing errors, detecting and possibly removing unwanted outliers and artifacts, and filtering features that do not appear to be scientifically or quantitatively useful. Treatment of missing observations should be done here as well; we recommend using multiple imputation methods \citep{Rubin2018, Carpenter2023}, as this will provide a naturally validatable approach. Additionally, feature engineering, i.e. the creation of new features based on the raw features available in the data, can be useful here to produce variables that have potentially much stronger associations or grouping structures in the data or which may be more pertinent scientifically compared to those in the raw feature set \citep{turner1999conceptual}.

In most cases, there exist multiple reasonable preprocessing strategies that can be used for any particular research question and data set. Also notably, it has been shown in many previous works that unsupervised techniques are extremely sensitive to changes in the preprocessing steps taken \citep{Crone2006, Denny2018, Zelaya2019, Alam2019, Schweinsberg2021}. Thus, the preprocessing steps taken should also be tested as part of the validation step discussed below. In particular, one should test the sensitivity of unsupervised discoveries to different preprocessing approaches in the workflow \citep{Yu2020}. This can be done by repeating the data preparation procedures with different, randomized choices of the preprocessing methods applied, e.g. changes in feature filtering decisions, hyperparameter selections, or outlier detection methods. Additionally, some approaches, such as multiple imputation techniques, are stochastic in nature and thus can be applied multiple times with different seeds to quantify sensitivity. 

\paragraph{Exploration.}

Alongside preprocessing, the data preparation step of the workflow also includes visualization and exploration of the data set. Exploratory data analysis is essential for gaining insight into the structures and patterns that exist in the data, including the existence of outliers, batch effects, and multivariate relationships between features. The knowledge gained from data exploration can be used in parallel with data preprocessing in order to inform necessary steps, such as whether additional outlier removal or data cleaning needs to be performed and what types of features may be of use to engineer or to filter out. The most important recommendation for the task of data exploration is to use as many different visualization formats as possible in order to fully explore the structure and relationships in the data. Different visualization methods and arrangements are best at displaying different aspects of the marginal and conditional distributions and multivariate relationships of features and thus can lead to different interpretations of the data \citep{Gleicher2011, Bresciani2015}. Further, different strategies may be required for visualizing high-dimensional data and relationships compared to low-dimensional data. Along the same lines, several types of unsupervised learning methods, particularly nonlinear dimension reduction and outlier detection with advanced machine learning techniques, create stochastic results, and thus repeated visualizations using these same techniques can help confirm the soundness of these explorations. We show an example of the benefits of using multiple visualization strategies as part of our case study in Section \ref{sec:case}.

\subsubsection{Modeling} In this step, the unsupervised learning models for analysis are selected and applied. Many resources currently exist which provide overviews of frequently used techniques for common unsupervised learning tasks, including clustering \citep{Kaufman2009}, dimension reduction \citep{Ghojogh2023}, network analysis \citep{Sucar2015}, and outlier detection \citep{Suri2019}, so we leave introductions and discussions on the specific techniques for each of these classes of unsupervised learning tasks to previous works. For choosing between different methods for a task, interpretability and concordance with addressing the scientific questions should be taken into account. For example, in the realm of dimension reduction, UMAP and tSNE will provide local interpretations of data structure while PCA will provide a global interpretation \citep{kobak2021initialization}, so care should be taken to align the appropriate method with the desired interpretation and scientific question.

Across all different types of unsupervised learning tasks, it is important to test multiple modeling techniques in the workflow for each individual task; this provides several advantages over solely applying a single method. Firstly, the trustworthiness of a finding can be measured by the frequency by which it appears across different modeling results. Previous studies have shown that, in general, if a discovery is true, then it should be apparent across different modeling techniques and hyperparameter selections \citep{rudin2024amazing, kuzilek2024rashomon,Yu2020}. Therefore, applying multiple techniques can help facilitate validation through providing a measure of robustness of findings based on the repeatability of each one across different applied methods and workflow settings. This technique for validation is especially useful in the context of unsupervised learning, as quantitative metrics for validating findings can otherwise be difficult compared to supervised learning, in which predictability of a target variable can be used. Secondly, in many cases, one can get different findings and results  from applying different methods addressing the same task \citep{sadoddin2007comparative}; methods for each task are not necessarily interchangeable. It can also be difficult to determine a priori which of a set of methods will produce the most useful results. Thus, applying multiple methods will increase the likelihood of generating data-driven discoveries through providing different modeling perspectives on the data, akin to the reasoning behind ensemble methods in supervised learning \citep{dietterich2000ensemble}. In addition, if one of the research goals is to produce a final trained model from multiple potential algorithms, then the application of multiple methods will generally require further model selection or model aggregation steps in order to produce the final model. General procedures for model selection will be discussed as part of validation in the following subsection.

For each unsupervised technique in the workflow, it is also important to test multiple hyperparameter settings and tune all hyperparameters in the validate stage, discussed subsequently. Several previous studies have shown that choosing different hyperparameter settings can yield vastly different results and conclusions \citep{Probst2019, Weerts2020, Arnold2024}.  While hyperparameter tuning is standard practice in supervised learning, this is less common in unsupervised learning where many instead select hyperparameters by looking at the data or based upon the concordance of results with prior knowledge.  These practices introduce subjective bias into the unsupervised workflow, however, and should hence be avoided.  Instead, we suggest testing multiple hyperparameter settings across each unsupervised method to ensure robustness of results to hyperparameter choices.  Additionally, testing multiple hyperparameters are tuning these in the validation stage, discussed next, is a critical component of selecting the final unsupervised model and reporting findings. 


\subsubsection{Validation} 

After selection of the models that will be used for analyzing the data, techniques and procedures for validating the results produced by said modeling procedures should be determined. Validation techniques are needed to assess reliability of results across all facets of the workflow. In particular, quantitative validation is essential for assessing the generalizability of results to new data and different decisions in modeling procedures, as well as promoting trust in data-driven discoveries \citep{Mayer1993, konig2007practical}. While human verification via literature review and qualitative comparisons can be helpful, it can be wrought with human biases and deduction of patterns that do not exist \citep{Parkavi2018, Sengupta2018, Sun2020}. Ideally, validation analyses should be performed via followup experiments, in which completely new data are collected and used to test whether the same findings are detected. However, in most cases collecting new data is costly in terms of time and material resources, and thus is not feasible. Thus, validation is instead typically performed using the data that is already available through creating a training-test split during data preparation, or by applying resampling strategies which simulate refitting models on new data sets \citep{NationalAcademies2019, meng2020reproducibility}. 

Below, we define common frameworks for validating unsupervised discoveries using available data as well as model-agnostic validation measures to assess the stability and generalizability of unsupervised findings. Notably, for many unsupervised approaches, standard model-agnostic methods for validation currently do not exist and this is an open area of research.  This occurs as unlike supervised learning, there is no ground-truth target to compare our findings against. However, when applicable, we introduce general validation methods that can be used across different methods for specific unsupervised learning tasks. We also note that the validation techniques to be used should be a critical consideration when planning the other parts of the workflow, including how the preparation and modeling methods selected can be harmonized with the validation plan as well as how to best communicate the uncertainty and reliability results from applying validation techniques. 

\paragraph{Validation Metrics}

One primary assessment for validation in unsupervised discovery is the \textbf{stability} of results, i.e. how similar or reliable the modeling results and interpretations of findings are with respect to arbitrary randomness and perturbations in data as well as differences in modeling decisions that can occur during preprocessing and modeling \citep{bousquet2002stability, yu2013stability, kleijnen1999validation,Yu2020, Belouafa2017}. Often, stability approaches utilize random perturbations of the observations, i.e. subsampling, bootstrapping, or random additions of noise, as well as different selections of hyperparameter settings. If a particular method produces stochastic outputs based on random initialization, e.g. UMAP or k-means clustering, using multiple iterations of this method can also be used to quantify stability. Metrics in the stability framework generally measure goodness-of-fit as a function of proportion of resamples in which individual results are repeated. Procedures for stability estimation have become well-established for certain specific unsupervised learning tasks; for example, in clustering, consensus clustering \citep{Monti2003, Goder2008} aggregates results from repeated clustering runs on data subsamples into a single metric that is weighted towards agreement of clustering assignments between each pair of observations across iterations. Similarly for graphical models, StARS \citep{Liu2010} measures stability via agreement of edge selections between pairs of features across model estimates from resampled data sets. In anomaly detection applications, rank stability methods that utilize the invariance of the magnitude of outlier measure have been proposed \citep{Perini2020}. Also, for nonlinear dimension reduction, previous work has proposed methods that quantify the variability in the embeddings across bootstrap samples \citep{Sun2023} or through the similarity in nearest neighbor structure with respect to the original data \citep{Xia2024}. In other unsupervised learning tasks, the resampling-based framework can still be used, but specific measures for evaluating the goodness-of-fit are not well-established and thus should be determined by the specific problem. 

Besides stability, another model-agnostic validation measure is \textbf{generalizability}, i.e. how well the modeling results hold if the analysis pipeline were to be rerun given new observations and if different modeling decisions are made given these new observations \citep{webb2005generalizability, Yarkoni2022}. This is of particular importance in assessing the applicability of discoveries to broader settings, such as different subpopulations or processes \citep{gobo2004sampling}. Measuring generalizability in the unsupervised learning paradigm is generally more difficult than measuring stability, as for many tasks the quantitative delineation of "good" as compared to "bad" generalizability is unclear. One potential general paradigm for measuring generalizability that may be useful for certain methods is the idea of using prediction validation with a supervised learning model and training-test split \citep{tibshirani2005cluster, Lange2004}. For example, in clustering tasks, clustering methods would be applied to both the training and test data sets to obtain estimated cluster labels; a supervised classification model is then fit using the training data set to predict the training cluster labels, and predictions of cluster labels are made using this model for the test data set. The generalizability of each clustering model result is then calculated as the concordance between the predictions from the classification model and the estimated cluster labels on the test set \citep{Handl2005}. Similar approaches to this have been proposed in recent work for specific unsupervised methods, including PCA \citep{Josse2012}, nonlinear dimension reduction \citep{Rostami2022}, and Gaussian graphical models \citep{Drton2007}, but this framework has not been widely established and is still an open area of research.

\paragraph{Model Selection \& Hyperparameter Tuning}

In the case where multiple techniques are applied in the modeling step or where multiple preprocessing pipeline decisions are plausible, the task of choosing a single "best" pipeline is often desired as part of the overall research objective.  Often, hyperparameter tuning is important as well, as many studies have shown that the choice of hyperparameters can have a large impact on the final results of data analysis pipelines in a wide variety of data contexts \citep{Schratz2019, Weerts2020, Soenen2021, Liao2022, Arnold2024}. Like with validation in general, model selection and hyperparameter tuning is less straightforward in the unsupervised learning framework due to the lack of a response variable for comparison. Here, both the stability-based and generalizability-based quantitative metrics mentioned above that are used for validating results can also be applied for this task; the final selected model is then the overall combination of preprocessing steps, model, and hyperparameter settings that produces the optimal level of the stability or generalizability metric \citep{Yu2020hyper}. For example, to select the number of clusters for a particular clustering method, the consensus clustering method mentioned above can be implemented across differing numbers of cluster centroids; a stability score threshold or elbow plot can then be used to select the optimal number of clusters required that meets an acceptable quantitative performance level. If multiple clustering methods have been used, this procedure can be applied for each individual clustering technique before selecting among these clustering methods, paired with their optimal number of clusters, to obtain the final model.

\subsubsection{Communication} 

Lastly, strategies for communication of ideas, results, and insights need to be determined for the workflow. The communication of outcomes and results is extremely important in collaborative work \citep{slade2023essential}, and thus should be a central focus of the planning of the workflow. Importantly, communication between data scientists and other research collaborators should occur before, during, and after analyses are performed with the workflow. During the planning stage, communication with collaborators is important to ensure that all stakeholders on the project are in alignment on the goals and objectives \citep{bennett2012collaboration}. Communication of methods and outcomes during data exploration and modeling will help ensure that results are correctly interpreted \citep{morrison2022challenges}. As modern unsupervised machine learning techniques are capable of generating findings that are not scientifically relevant in context \citep{allen2023interpretable, Ashtari2023}, results should be evaluated by researchers with domain-specific knowledge in order to ensure outcomes will produce useful insights in context. Lastly, effective communication of findings at the end of the pipeline is important for scientific validation of results, e.g. determining which discoveries are important. These discussions can also serve as a catalyst for formulating new hypotheses for follow-up analyses \citep{elnaqa2018machine}. For general communication with collaborators, it is important to have simple, clear explanations of statistical methods and contextually relevant interpretations of findings and the insights that can be derived from them \citep{clare2017communicating}. Also, in addition to the raw findings and results, the communication plan should include a discussion uncertainty and limitations of work in order to provide the audience with a gauge on the reliability of results.

Effective data visualizations are also an extremely important part of understanding data sets and results as part of any data science workflow \citep{islam2019overview, unwin2020why}. However, this is especially important for unsupervised learning workflows, as quantitative assessments of results are often not available due to lack of ground truth for comparison for many tasks. Thus, we recommend that visualizations be created to study outputs at nearly every step of the pipeline. Additionally, we suggest using multiple visualizations that vary the aesthetic mappings to features of the graph as well as the type of graphic in order to maximize the information gleaned from the data and modeling and to minimize the probability of biases or misinterpretations that arise from only using a single visualization \citep{roberts1998encouraging, roberts2000multiple}. The creation of interactive visualization tools can also be extremely valuable for understanding interactions between features in large, high-dimensional data sets and for allowing audiences to explore data and results \citep{ward2010interactive, janvrin2014making}. Many resources are available that outline the core concepts of creating statistical graphics, e.g. \citep{wickham2017r, evergreen2019effective, healy2024data}; we recommend learning and applying these in order to produce effective visualizations for data exploration and communicating results. 

Another important facet of communication is reproducibility. In unsupervised learning (and more broadly across all data science problems), the topic of reproducibility has been a preeminent topic of discussion \citep{Willis2021, Fineberg2020}. Showing reproducibility and reliability of the unsupervised learning workflow and the discoveries produced is paramount for promoting trust in the modeling results \citep{meng2020reproducibility, Stodden2020}. Generally, there are several different facets of reproducibility and reliability that need to be shown. In terms of computational reproducibility, the same code, deterministic algorithms, and data should be able to produce the exact same results, and the same code with stochastic algorithms, e.g. different simulation runs or random initializations, should produce similar results. To this end, all computational tools and procedures should be made available. For assessing statistical replicability, all modeling decisions and justifications for all steps of the analysis should be documented and considered as part of the final communication plan. 


\subsection{Best Practices for Unsupervised Discovery}\label{sec:bestprac}

Here, we summarize the best practices across the different facets of the unsupervised workflow described above. The recommendations we discuss below reiterate the core concepts of building an unsupervised learning workflow that should always be considered to produce more reliable unsupervised scientific discoveries.

\paragraph{Plan Ahead.} 

All aspects of the unsupervised learning workflow should be planned out before any actual analysis occurs. While changes may need to be made afterwards due to unforeseen circumstances, the lack of planning can often lead to repeating  previous work. Common examples of this include having insufficient data, specifying scientific questions not conducive to objective validation, or failure to create a training-test split for validation a priori.

\vspace{-0.25cm}\paragraph{Communicate Frequently.}  

Communication between the data scientist and other members of the project should occur regularly throughout the course of the analysis, starting from the planning stage all the way until results are finalized. This will help ensure results are scientifically meaningful and novel.

\vspace{-0.25cm} \paragraph{Visualize, Visualize, Visualize.} 

Data visualization is the best way to help understand the structures and associations within a dataset. Visualizations should be used at each step in the workflow to assess the results that are produced by each method applied. Also, using multiple visualizations at each step will aid in gaining a comprehensive understanding of the data.

\vspace{-0.25cm} \paragraph{Use Multiple Techniques.} 

For each step in the exploration and modeling phases, using multiple techniques can assist in generating discoveries through providing different insights on the dataset. Thus, the workflow should include more than one approach for each step in order to maximize the potential for relevant findings as well as assess the robustness of findings to arbitrary modeling choices.

\vspace{-0.25cm}\paragraph{Demonstrate Reproducibility, Replicability, and Reliability.} 

In addition to statistical validation, the reproducibility, replicability, and reliability of results need to be addressed as well. Code should be documented and provided to show computational reproducibility. Decisions made in terms of preprocessing and modeling choices should be documented and communicated as part of the results in order to appraise replicability. Lastly, validation of all steps of the workflow should be performed to assess reliability.

\vspace{-0.25cm}\paragraph{Validate, Validate, Validate.} 

All results created from the workflow should be validated in order to provide a gauge for the trustworthiness and reliability of all data-driven discoveries; this includes validating both the raw outcomes from the quantitative approaches at each stage as well as the consequent scientific discoveries produced by these outcomes. Most importantly, reliability needs to be assessed across all different aspects of the workflow; this includes measuring the effect of preprocessing choices, hyperparameter tuning, and model selection. Additionally, if possible, both stability and generalizability of results should be considered in validation.

\section{Case Study: Finding Common Origins of Milky Way Stars} \label{sec:case}

To help illustrate these best practices, we next provide a case study, investigating the shared origins (or clusters) of Milky Way stars based upon their chemical composition. In this section, we begin by introducing the scientific question and data. We then detail our analysis plan. Finally, we present the results of our investigation, proceeding from an exploratory data analysis through a stability-driven clustering analysis and concluding with a rigorous validation of our identified clusters. Through this case study, we illustrate how the benefits of a carefully-designed workflow for unsupervised learning can advance scientific discovery. 

\subsection{Scientific Question and Data}
The emergence of large spectroscopic surveys of the Milky Way has led to significant interest in studying the chemical origins of the Galaxy's formation. The Apache Point Observatory Galactic Evolution Experiment (APOGEE) is a high-resolution, near infrared survey of stars comprising the disk, or the primary area of the Milky Way's stellar mass \citep{prieto2008apogee}. Features include chemical abundances of 19 key elements in star formation, radial velocities, and meta information for each star such as surface gravity and effective temperature. Researchers are often interested in using the APOGEE data to explore the stellar origins of the Milky Way and specifically how stars were formed or evolved chemodynamically over large periods of time \citep{Baron_Poznanski_2016, casamiquela2021possibility, Donor_Frinchaboy_Cunha_O’Connell_Prieto_Almeida_Anders_Beaton_Bizyaev_Brownstein_et, Fraix-Burnet_Bouveyron_Moultaka, Hawkins_Jofré_Masseron_Gilmore_2015, Hayden_Bovy_Holtzman_Nidever_Bird_Weinberg_Andrews_Majewski_Allende}. Existing observations show that in the process of forming a stellar body, parent molecular clouds can produce hundreds of stars in a single burst \citep{krumholz2014star}, but due to astronomical dynamics, these shared origins are challenging to find. In this case study, we hence aim to cluster stars \textit{chemically} to identify shared origins and trace the history of the Milky Way.

Globular clusters (GCs) are an ideal population to begin studying the origins of Milky Way development. GCs are sites of dense star formation, comprised of $10^5-10^7$ stars formed at the same time, and have been shown to be critical building blocks of stellar bodies such as halos, discs, and spheroids \citep{tremaine1975formation, martell2011building, debattista2023imprint, kane2025ones}, with their full impact remaining an open research question in the field \citep{rusta2024linking}. GCs exhibit specific chemical abundance trends \citep{gratton2004abundance, milone2022multiple, marino2023metallicity}, making them a prime target for assessing clustering to find shared origins of the stars. Still, while many stars in the APOGEE survey have been assigned to GCs based on current proximity, researchers are also interested in identifying deeper shared origins of possibly distant bodies.  There exists a wide body of literature, which here we only refer to a small subset, aiming to do exactly that, not only in GCs but in other stellar populations within APOGEE \citep{ratcliffe2020tracing, pat2022reconstructing, kane2025ones, Baron_Poznanski_2016, Hawkins_Jofré_Masseron_Gilmore_2015, legnardi2022constraining, belokurov2023nitrogen}. In short, this literature focuses on grouping GCs in order to find their common origins; what progenitors they share, how the descendants are redistributed around the galaxy, and how they can be distinguished. Through these groupings, astronomers seek to characterize the underlying evolution in the age and metallicity of the Milky Way to trace its assembly.  Translating this scientific question into a validatable data science goal, our objective is to further group GCs based on their chemical signatures to find reliable, robust, and scientifically interpretable clusters of stars.  

\subsection{Plan Ahead}

Before undertaking this data analysis, it is crucial to plan ahead in all stages of the scientific workflow. This planning ahead includes both specifying various analytical choices to be made and outlining the steps in the overall workflow.

\subsubsection{Data Preprocessing and Modeling Choices}

To begin, there are numerous human judgment calls or choices that inevitably must be made throughout the unsupervised workflow. In particular, though the pipeline of data processing from raw infrared spectra data to chemical tagging is considered to be standard and well-defined \citep{nidever2015data}, there are many different, but equally-reasonable data preprocessing choices that have been explored in the previous scientific literature, including different quality control filtering thresholds, data imputation procedures, and subsets of elements that are included and scientifically useful for clustering \citep{ting2022many, price2020strong, casamiquela2021possibility, chen2018chemodynamical}. 
Within the modeling stage, different studies have used different clustering approaches, ranging from density-based approaches such as HDBSCAN to K-Means, Gaussian Mixture Models, and deep learning-based approaches \citep{casamiquela2021possibility, garcia2018machine, chen2018chemodynamical,ratcliffe2020tracing, pagnini2025abundance,bialopetravivcius2020deriving, cavallo2023parameter, berni2024searching}. Each of these studies finds a different number of clusters, with several concluding the lack of feasibility of clustering this data with confidence \citep{ratcliffe2020tracing, casamiquela2021possibility}. Efforts to statistically validate these clusters have also been limited, with studies rarely comparing the stability and generalizability of these methods, or their sensitivity to different preprocessing choices. Similar difficulties arise with dimension reduction, with some studies using tSNE \citep{anders2018dissecting}, some with UMAP \citep{casamiquela2021possibility}, and some using PCA \citep{ratcliffe2020tracing} to visualize and analyze the APOGEE data.

For our case study, we detail the various data preprocessing choices, dimension reduction methods (with hyperparameters), and clustering methods (with hyperparameters) under consideration in Table~\ref{tab:choices}. Briefly, we consider different choices of quality control filters, sets of elemental abundance features used in the analysis (either 7, 11, or 19 features, each standardized to have mean 0 and standard deviation 1), data imputation methods (using mean-imputation or random forest-based imputation \citep{stekhoven2012missforest}), dimension reduction methods prior to clustering (none, PCA \citep{hotelling1933analysis}, tSNE \citep{van2008visualizing}, and UMAP \citep{mcinnes2018umap}, each with various hyperparameters), clustering methods (spectral clustering \citep{alpert1999spectral}, K-means \citep{arthur2006k}, and hierarchical clustering \citep{mojena1977hierarchical}, each with various hyperparameters), and finally the total number of clusters ($k = 2, \ldots, 30$).

\begin{table}[t]
\scriptsize
\begin{tabular}{@{}lll@{}}
\toprule
& \textbf{Data Preprocessing/} & \multirow{2}{*}{\textbf{Parameter Choices}} \\
& \textbf{Method} & \\
\midrule
\parbox[t]{3mm}{\multirow{10}{*}{\rotatebox[origin=c]{90}{\textbf{Data Preprocessing}}}}
& \multirow[t]{5}{*}{Quality Control Filter}
& S/N Threshold = 30, 50, 70$^*$, 90, 110, 130, 150 \\
&
& $T_{eff}$ Width = 500, 1000$^*$, 1500, 2000 \\
& 
& Log Gravity Threshold = 3.0, 3.3, 3.6$^*$, 3.9, 4.2 \\
& 
& VB Threshold = 0.5, 0.7, 0.9$^*$, 0.99 \\
&
& STARFLAG = 0 \\
\cmidrule(l){2-3}
& \begin{tabular}[c]{@{}l@{}} Abundance Features \\ \\ \\ \end{tabular}
& \begin{tabular}[c]{@{}l@{}}
    (i) 7 features: $FE_{H}, MG_{FE}, O_{FE}, SI_{FE}, CA_{FE}, NI_{FE}, AL_{FE}$,\\ 
    (ii) 11 features: (i) plus $C_{FE}, MN_{FE}, N_{FE}, K_{FE}$, or\\ 
    (iii) 19 features: (ii) plus 
$CI_{FE},\,NA_{FE},\,S_{FE},\,TI_{FE},\,TIII_{FE},V_{FE},\,CR_{FE},\,CO_{FE}$
\end{tabular} \\
\cmidrule(l){2-3}
& Data Imputation 
& Mean imputation or random forest-based imputation \\
\midrule
\parbox[t]{3mm}{\multirow{8}{*}{\rotatebox[origin=c]{90}{\begin{tabular}[c]{@{}c@{}}\textbf{Dimension}\\\textbf{Reduction}\end{tabular}}}}
& No Dimension Reduction & NA \\
\cmidrule(l){2-3}
& PCA & NA \\
\cmidrule(l){2-3}
& \multirow[t]{2}{*}{tSNE} 
& Number of dimensions = 2 \\
& & Perplexity = 10, 30, 60, 100, 300 \\
\cmidrule(l){2-3}
& \multirow[t]{2}{*}{UMAP} 
& Number of dimensions = 2 \\
& & Number of neighbors = 10, 30, 60, 100, 300 \\
\midrule
\parbox[t]{3mm}{\multirow{5.5}{*}{\rotatebox[origin=c]{90}{\begin{tabular}[c]{@{}c@{}}\textbf{Clustering}\\ ($k = 2, \ldots, 30$)\end{tabular}}}}
& K-means & Initialization = K-means++ \\
\cmidrule(l){2-3}
& \multirow[t]{2}{*}{Hierarchical Clustering}
& Distance = Euclidean \\
& & Linkage = complete or Ward's \\
\cmidrule(l){2-3}
& \multirow[t]{2}{*}{Spectral Clustering} 
& Affinity = nearest neighbors graph \\
& & Number of neighbors = 5, 30, 60, 100 \\
\midrule
\end{tabular}
\caption{Data preprocessing and modeling choices used in the APOGEE case study. Quality control parameters with astericks were used as the default/baseline quality control parameters.}
\label{tab:choices}
\end{table}

\subsubsection{Workflow Plan}

We next outline our workflow, organized into three parts: (1) data preparation and exploration, (2) modeling and validation, and (3) interpretation and communication.

\paragraph{Data Preparation and Exploration}

We leverage data from the APOGEE DR17 value-added catalog of GCs \citep{schiavon2024apogee}, and as a baseline, we first follow the quality control filtering steps from \citep{pagnini2025abundance} (details in the Interactive Supplement (IS)\footnote{\url{https://dataslingers.github.io/unsupervised-workflow-astro/}} Section 2). This results in a processed dataset with 3,286 stars and 19 chemical abundance features. With these 3,286 observations, we perform a randomized 80/20 train/test split to obtain a training set of 2,628 observations and a test set of 658 observations. Alternative data preprocessing pipelines will also be considered and discussed in subsequent sections.

We then conduct a brief exploratory data analysis, examining the spatial distribution of GCs, the distribution of abundances, and the pairwise relationships between different elemental abundances.
We further explore various dimension reduction visualizations of the abundance data using PCA \citep{hotelling1933analysis}, tSNE \citep{van2008visualizing}, and UMAP \citep{mcinnes2018umap}. For tSNE and UMAP, we examine different hyperparameter choices of the number of nearest neighbors (known as perplexity in tSNE) used to construct the manifolds. Intuitively, this hyperparameter controls the balance between preserving local versus global structure in the learnt representations. To tune this hyperparameter and select the best dimension reduction method, we use the neighborhood retention metric, which measures the proportion of the $k$ nearest neighbors maintained from the original high-dimensional space to the low-dimensional embedding \citep{xia2024statistical, liu2024assessing}. We assess this metric across a range of neighborhood sizes $k$. A ``good'' dimension reduction method and hyperparameter exhibits high neighbor retention across a wide range of neighborhood sizes. 

\paragraph{Modeling and Validation}

To next identify the optimal clustering of stars, we introduce our clustering and validation workflow.
Given the many reasonable data preprocessing and modeling choices, our approach is heavily driven by the philosophy that our scientific conclusions should be stable and generalizable regardless of which specific analytical choices are made. This is similar in spirit to the core principles of the PCS framework for veridical data science \citep{Yu2020}, which emphasize that scientific conclusions should, at a minimum, be stable across arbitrary human judgment calls made throughout the data science life cycle. Guided by this philosophy, we summarize our approach as follows:

\begin{adjustwidth}{0.5cm}{}
\vspace{0.5em}
\noindent\textbf{(a) Stability-driven model building.} 
Given training data $X$, a set of clustering methods of interest $\mathcal{M}$, a set of reasonable data preprocessing pipelines $\mathcal{G}$, and a range of number of clusters $\mathcal{K}$, we first select the clustering model $m^* \in \mathcal{M}$ and the number of clusters $k^* \in \mathcal{K}$ that yield the most stable (i.e., similar) clustering results, when fitted using different subsamples of $X$ and different preprocessing pipelines $g(X)$ for all $g \in \mathcal{G}$. Here, we measure the overall stability (or similarity) between two sets of clusters using the Adjusted Rand Index (ARI) \citep{rand1971objective}. Pseudo-code for this model selection step is provided in Algorithm~\ref{alg:model_explorer}.

This model selection step is similar to the Model Explorer algorithm from \citep{ben2001stability}, but while the Model Explorer only assesses cluster stability across random subsamples of the data, we extend this to include the stability across different preprocessing pipelines. This is crucial, as scientifically-relevant clusters should not depend on arbitrary data preprocessing choices.

To then obtain the final cluster labels for the training data, we use consensus clustering \citep{monti2003consensus} to aggregate the clusters learned using the most stable model $m^*$ and number of clusters $k^*$, applied to each preprocessed version of the data $g(X)$ for each $g \in \mathcal{G}$. Here, we report the co-clustering membership matrix, which captures the frequency of times each pair of stars appeared in the same cluster across the different clustering runs using different data preprocessing pipelines $\mathcal{G}$. From this co-clustering matrix, we further compute the local stability of each star's cluster membership, defined as the average co-clustering frequency across all other stars in the same cluster. A higher local stability indicates that this star is stably clustered with its neighboring stars regardless of the data preprocessing choices.

\vspace{0.5em}
\noindent\textbf{(b) Generalizability and stability validation.} 
To validate the consensus training clusters, we assess (i) the overall stability had alternative preprocessing choices (not considered in $\mathcal{G}$ due to computational feasibility) been performed, and (ii) the cluster generalizability using a held-out test set. More specifically, to measure (i), we compute the ARI between the previously learned clusters and the clusters that would have been obtained had alternative preprocessing choices been made. To measure (ii), we follow the idea of cluster predictability \citep{tibshirani2005cluster, Lange2004}: we apply the tuned clustering pipeline to the training and test sets separately, train a random forest (RF) to predict the cluster labels on the training set, and use this RF to predict the cluster labels on the test set. The overall cluster generalizability is then computed as the ARI between the RF-predicted cluster labels and the test cluster labels. However, since some clusters might be more generalizable than other clusters, we further investigate a measure of the local, or cluster-wise, generalizability by computing the precision between the RF-predicted cluster labels and the test cluster labels \textit{per cluster}.
    
If the clusters validate, we finally re-fit the tuned clustering pipeline on the full training and test data to obtain the final clusters and interpret them scientifically.

\end{adjustwidth}

\begin{algorithm}[tb]
\caption{Clustering Model Selection via Stability}\label{alg:model_explorer}
\begin{algorithmic}[1]
    \Require Training data $X$, set of preprocessing pipelines $\mathcal{G}$, set of clustering methods $\mathcal{M}$, set of cluster numbers $\mathcal{K}$, resampling iterations $B$, and subsampling proportion $\pi$.
    \For{$k \in \mathcal{K}$}
        \For{$m \in \mathcal{M}$}
            \For{$b = 1, \ldots, B$}
                \State Randomly sample a subset of observations: $sub_1 = \text{subsample}(X, \pi)$.
                \State Randomly sample a subset of observations: $sub_2 = \text{subsample}(X, \pi)$.
                \State Randomly choose two data preprocessing pipelines: $g_1, g_2 \in \mathcal{G}$.
                \State Apply clustering method $m$ on $g_1(X[sub_1])$ to obtain $k$ clusters: $C_1$.
                \State Apply clustering method $m$ on $g_2(X[sub_2])$ to obtain $k$ clusters: $C_2$.
                \State Compute stability score via Adjusted Rand Index (ARI) on overlapping samples:
                $$S(k, m, b) = ARI(C_1[sub_1 \cap sub_2], C_2[sub_1 \cap sub_2]).$$
            \EndFor
        \EndFor
    \EndFor
    \State Choose clustering method $m^* \in \mathcal{M}$ and number of clusters $k^* \in \mathcal{K}$ with highest stability:
    $$m^*, k^* = \argmax_{k \in \mathcal{K}, m \in \mathcal{M}} \;\; \frac{1}{B}\sum_{b = 1}^B S(k, m, b).$$
\end{algorithmic}
\end{algorithm}

\paragraph{Interpretation and Communication}

We report both statistical and scientific findings from the final learned clusters. This includes stability and generalizability metrics as well as scientific insights stemming from the chemical abundance signatures of each identified cluster. In particular, through these elemental signatures, we compare our discovered clusters to existing observations in the literature and assess whether our discovered clusters unearth new groupings of globular clusters as compared to the current understanding of star formation.

\subsection{Results}

\begin{figure}[p]
    \centering
    \includegraphics[width=1\linewidth]{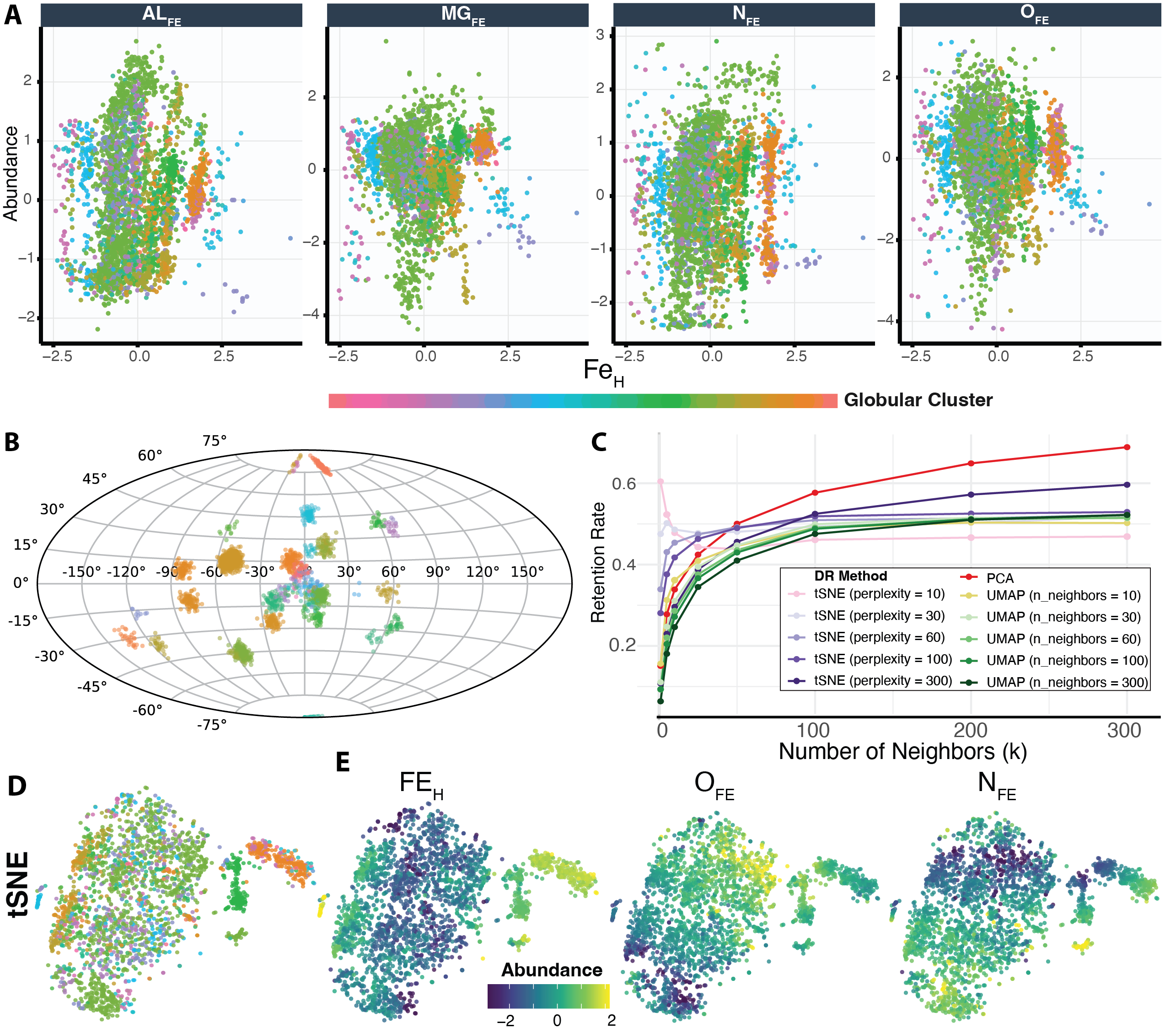}
    \caption{Exploratory Data Analysis of APOGEE DR17 Globular Clusters (GCs). \textbf{A:} Distributions of key chemical abundances across $FE_H$. \textbf{B:} GCs visualized in galactic coordinate space. As expected, GCs are spatially grouped. \textbf{C:} GC neighborhood retention for different dimension reduction (DR) methods and hyperparameters across different neighborhood sizes. Across PCA, five hyperparameter settings of tSNE, and five hyperparameter settings of UMAP, we computed how many of the $k$ nearest neighbors are maintained from the original chemical abundance space to the 2-dimensional embedding. tSNE always maintains a higher neighborhood retention rate than UMAP across all hyperparameters and sizes of neighborhoods. To balance local information, we select tSNE with perplexity=100 for further visualizations. \textbf{D:} tSNE with perplexity=100 embedding of GC labels. \textbf{E:} Key chemical abundance distributions along the tSNE (perplexity=100) embedding. From these, we can observe iron-rich, younger GCs in the leftmost plot, oxygen-rich GCs in the middle plot, and nitrogen-rich GCs on the right.}
    \label{fig:fig1}
\end{figure}

\paragraph{Exploration} Following our plan, we first carry out an exploratory data analysis on the training APOGEE DR17 data (preprocessed using the baseline quality control filters, 11 elemental abundance features, and mean imputation), seen in Figure \ref{fig:fig1}. In particular, since $FE_H$ is a commonly agreed upon marker of star age \citep{anders2023spectroscopic, schiavon2024apogee}, we hone in on pairiwse relationships between $FE_H$ and four other key abundances ($AL_{FE}, MG_{FE}, N_{FE},$ and $O_{FE}$) in Figure~\ref{fig:fig1}A. We observe that the majority of the separation between GCs seems to be occurring in the $FE_H$ axis as compared to the other elements. In Figure \ref{fig:fig1}B, we also examine the GCs in galactic coordinate space. As expected, stars share GCs with their spatial neighbors. 

We next explore various dimension reduction techniques. To identify appropriate hyperparameters for these methods, we evaluate the neighbor retention metric for various dimension reduction methods and hyperparameters in Figure \ref{fig:fig1}C. Interestingly, tSNE almost always uniformly demonstrates higher retention than UMAP. For this reason, we drop UMAP from consideration in all subsequent analyses. We also note that tSNE with perplexity=30 and perplexity=100 exhibit high retention across both small (i.e., local) and large (i.e., global) neighborhood sizes and are thus deemed appropriate for subsequent steps (e.g., for visualization and as input to clustering methods), alongside PCA which expectedly has the highest retention globally. Having selected and tuned these dimension reduction techniques, in Figure \ref{fig:fig1}D, we visualize the GC labels on the tSNE (perplexity=100) embedding, and in \ref{fig:fig1}E, we overlay several key chemical abundances ($FE_H, O_{FE}, N_{FE}$). We can see that the right arm of the tSNE embedding is primarily iron-rich stars, a marker of younger stars \citep{chiao2015young}. We provide further exploratory plots and dimension reduction visualizations in IS.3-4.

\paragraph{Modeling and Validation} Using our training data, we next search over different clustering methods to find the most stable clusters of stars using Algorithm~\ref{alg:model_explorer} and summarize the stability results in Figure~\ref{fig:fig2}A. We shortlist two promising methods, spectral clustering (n\_neighbors = 60) with $k=2$ clusters and K-means clustering with $k=8$ clusters, which demonstrate high stability across both subsamples of the data and different preprocessing pipelines. 

First, exploring the spectral clustering (n\_neighbors = 60) with $k=2$ clusters, we visualize the consensus clustering matrix in Figure~\ref{fig:fig2}B alongside the tSNE (perplexity=100)  embeddings, overlayed with the cluster membership labels and the local stability of each star's cluster membership in Figure \ref{fig:fig2}C-D, respectively. We observe that the two clusters are extremely stable and exhibit high local stability values across almost all stars. However, despite yielding the highest stability in Figure~\ref{fig:fig2}A, these two clusters largely recapitulate well-established iron-rich versus iron-poor splits (Figure~\ref{fig:fig2}E), characterizing younger versus older stars \citep{chiao2015young, chiappini2015young,chiao2019gauge}. 

Seeking new insights beyond these known divisions, we instead shift our interest towards the eight clusters generated by K-means (Figure~\ref{fig:fig2}F-H) with the hopes of uncovering novel and stable groupings of GCs. Notably, several clusters - 1, 2, and 8 - show strong co-clustering membership, but others - 3, 7 - show high variability in co-clustering membership across pipelines. These cluster-specific insights perhaps suggest that while not all clusters in the K-means ($k=8$ clusters) model should be trusted or interpreted, there are several clusters that are highly stable and may give rise to more reliable scientific conclusions.



\paragraph{Interpretation and Communication}
Proceeding with the selected K-means clustering model with $k = 8$ clusters, we finally refit the consensus clustering pipeline on the full datasets with no sample splitting. In Figure~\ref{fig:fig4}A, we highlight these final learned clusters on the tSNE (perplexity=100) visualization. In \ref{fig:fig4}B, we evaluate the robustness of this final clustering determination with respect to changing astronomical parameters in the quality control filtering step. Instead of the original thresholds chosen for S/N (70) and log of surface gravity (3.6), we sweep over alternative choices for these thresholds and confirm that the final clusters are stable (or similar) regardless of these arbitrary filtering thresholds (see IS.6 for additional stability plots). In Figure~\ref{fig:fig4}C, we assess the local generalizability across the clusters in our held-out test set, averaged across imputation methods and datasets (see IS.6 for more detail). In \ref{fig:fig4}D, we can directly compare this with the stability of individual clusters. We observe that cluster 2 has the highest generalizability and the highest stability, indicating that it is both stable to data perturbations and generalizes extremely well to new data. 
One possible reason for this strong performance of cluster 2, in addition to cluster 1, is they are heavily composed of large GCs, which are entirely self-contained within the cluster (Figure~\ref{fig:fig4}E). Specifically, NGC6121 makes up 55\% of cluster 1, and 98.8\% of the GC is contained within it. Similarly, NGC0104 makes up 70\% of cluster 2 and is wholly contained within it. 

Due to the strong generalizability and stability performance of clusters 1, 2, 6, and 8 in \ref{fig:fig4}C and D, we determine these four groupings as the most robust, and hence suitable for scientific interpretation and communication to collaborators. In Figure~\ref{fig:fig4}F, we examine the distributions of seven key chemical abundances over these groups. Though we briefly describe the clusters and their relevance here, we refer readers to the rich body of astronomical literature on interpreting clusters of GCs, such as but not limited to \citep{schiavon2017chemical, ratcliffe2020tracing, casamiquela2021possibility, fernandez2020aluminium, belokurov2023nitrogen}. The unique chemical abundance of groups of stars traces the formation and assembly of these stars from possibly common progenitors. Cluster 2 is characterized by heavy iron-richness but low aluminum, indicative of a cluster of iron-rich metal-poor (IRMP) stars, potentially tied to specific supernovae as formation systems \citep{reggiani2023iron}. Cluster 6, with low abundances of carbon, magnesium, oxygen, and silicon, is more challenging to identify. It is possibly another cluster of metal-poor stars \citep{fernandez2020aluminium}, however unlike cluster 8, it is not iron-poor. Clusters 1 and 8 are high in nitrogen, and could be considered a set of nitrogen-rich stars (NRS), particularly cluster 8 which is also high in key marker aluminum and with significantly low abundance in $FE_H$ (see \citep{schiavon2017chemical}). These older, metal-poor stars have been shown to be particularly important in the early formation of the inner galaxy \citep{fernandez2019chemodynamics, fernandez2019discovery, fernandez2020dynamical}, and cluster 8 in particular exhibits a known anti-correlation between high aluminum abundance and low magnesium abundance \citep{baeza2022apogee}. While promising, we emphasize further research is needed to scientifically validate these clusters, ideally with a collaborating astronomer. Further research into clustering astronomical survey data may seek to integrate other commonly used value-added catalogs from APOGEE such as annotated age labels \citep{sanders2018isochrone}, other spectroscopic sky surveys such as GALAH \citep{de2015galah, buder2024galah} and Gaia \citep{gilmore2012gaia, gaia2023gaia}, or commonly used simulation studies for the evolution of the Milky Way \citep{mackereth2018origin, clarke2019imprint}.

The robustness of our results and pipeline underscores the importance and benefit of a strong workflow for scientific discovery in unsupervised learning. Though the chemical abundance space of stars in the Milky Way has been established as a continuous spectrum of age and metallicity \citep{ratcliffe2020tracing, bovy2016chemical, bovy2012milky}, our workflow has produced reliable and reproducible groupings of stellar abundances, anchoring data-driven insights into chemical formations of our galaxy. The need to cluster along a continuum is by no means unique to stellar astrophysics. Across disciplines, from cell cycle dynamics in cancer \citep{zikry2024cell} to agricultural soils \citep{heil2019advantages}, researchers frequently wish to partition and typify data along continuous spectra. Through this case study, we have demonstrated the value of a workflow with rigorous assessments of both local and global stability and generalizability to ensure reproducible results.

\paragraph{Data Availability}
For all code, documentation, justification of preprocessing and modeling choices, and more in-depth validation analyses and results, please see our Interactive Supplement at \url{https://dataslingers.github.io/unsupervised-workflow-astro/} and GitHub repository \url{https://github.com/DataSlingers/unsupervised-workflow-astro}. Our processed APOGEE globular cluster data with train-test splits and all results files are also made available at \url{https://zenodo.org/records/15565719}.

\section{Discussion \& Open Research} \label{sec:dis}

In this paper, we have provided a general, model-agnostic workflow for generating unsupervised data-driven discoveries as well as practical recommendations on best practices when designing and executing data analysis pipelines using unsupervised learning techniques. However, there are also several specific topics and questions that we have been unable to definitively address, as they are still areas of open research and thus do not have a well-established solution. The most outstanding of these issues is in the area of model-agnostic validation and uncertainty quantification techniques. While general approaches have been established for evaluating the stability and generalizability of clustering estimates \citep{monti2003consensus, roth2002resampling, tibshirani2005cluster}, model-agnostic methods for validation in other unsupervised learning tasks are currently open unsolved problems. Similarly, the concept of quantitative measures for comparing uncertainty levels between different models applied for the same unsupervised learning task is a difficult problem to address, as the notion of prediction errors from supervised learning analyses does not exist, and thus little work has been done in this area. Another area of open research involves methods for synthesizing results found from multiple analytical approaches, hyperparameter settings, or modeling decisions, analogous to ideas from ensemble learning in the supervised learning paradigm. The problem of feature importance is also presently an unsolved problem, as most methods in unsupervised learning do not provide parameter estimates which can be mapped to a measure of feature importance or a target to assess significance. Though we currently do not have clear recommendations for these topics, developments of methods for these concepts in the realm of unsupervised learning are a burgeoning area of interest in statistics, machine learning, and AI; thus these present unaddressed topics may have more definitive solutions in the near future. 

In summary, unsupervised learning methods have been important for driving research and creating new data-driven discoveries in the sciences and beyond, but guidance on how to create reliable workflows for unsupervised discovery is currently limited. Through the establishment of a general workflow for unsupervised learning, we aim to provide data scientists and interdisciplinary researchers with a framework for developing unsupervised analysis pipelines in new fields, as well as to promote reproducible and reliable data-driven discoveries. Development of generalized, modal-agnostic approaches for validation, uncertainty quantification, and post-hoc analysis of unsupervised learning methods will be an important area of future research for ensuring further reliability of unsupervised discoveries.

\begin{figure}[p]
    \centering
    \includegraphics[width=0.9\linewidth]{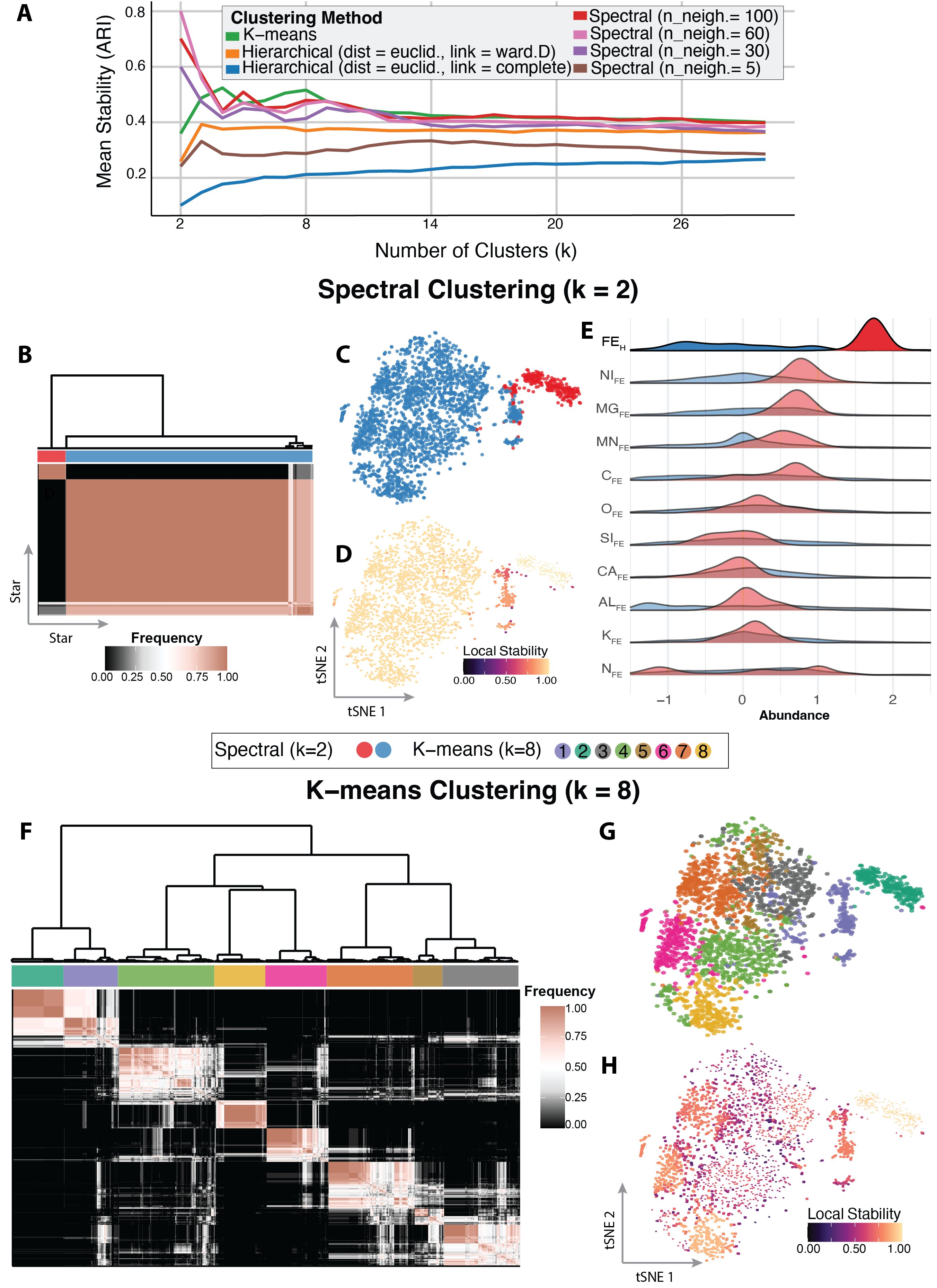}
    \caption{Stability-driven clustering model selection. \textbf{A:} Stability assessment of clustering methods on a training set of APOGEE globular clusters (GCs) using Algorithm~\ref{alg:model_explorer}. Aggregated over imputation methods and datasets (see Table~\ref{tab:choices}), we compute the mean stability for each clustering method. Spectral clustering with 60 nearest neighbors at $k=2$ has the highest overall mean stability, followed by K-Means with $k=8$. Hence, we explore these two clustering results on the training set.}
    \label{fig:fig2}
\end{figure}
\makeatletter
  \setlength{\OLDfptop}{\@fptop}
  \setlength{\@fptop}{0pt}
\makeatother
\begin{figure}[p]\ContinuedFloat
  \caption[]{(Previous page.) \textbf{B:} Consensus clustering matrix across imputation methods and datasets for the two cluster spectral solution (n\_neighbors=60, $k=2$). Each entry $(i,j)$ is the fraction of times stars $i$ and $j$ are in the same cluster across every run of the full clustering pipeline: imputation methods, feature sets, and DR methods (see Table~\ref{tab:choices}). \textbf{C:} tSNE (with perplexity=100) embeddings of the medium mean-imputed dataset, overlaid with consensus cluster labels. \textbf{D:} Local stability of the spectral clustering result overlaid on the tSNE embedding. Local stability for each star is computed as its average co-membership rate with its own cluster across the clustering pipeline. \textbf{E:} Chemical abundances of the two spectral clusters, sorted by magnitude of difference. Though other elements are used to group these stars, these clusters are predominantly being separated along the $FE_H$ axis, an already known distinguishing factor of young and old stellar bodies. This iron-rich population on the right arm of the tSNE embedding (panel C) is highly stable (panels B, D). Though exhibiting the highest stability, we instead choose to explore the K-means result with $k=8$ in the aims of uncovering more granular groupings of GCs rather than recapitulating already known distinctions between iron-rich younger stars and older stars. \textbf{F:} Consensus matrix for the eight clusters from K-means. \textbf{G:} tSNE embedding of the medium mean-imputed dataset overlaid with the eight clusters from K-means. \textbf{H:} Local stability of each of the eight K-means clusters. As observed in the consensus matrix (panel F), there are more unstable clusters as compared to the spectral clustering results (panels B-E). For example, clusters 3 and 7 exhibit low stability, first visualized in the light entries of the consensus matrix within their respective blocks, then overlaid in panel H. The iron-rich cluster 8 is the most stable.}
\end{figure} 

\newpage
\begin{figure}[p]
    \centering
    \includegraphics[width=0.9\linewidth]{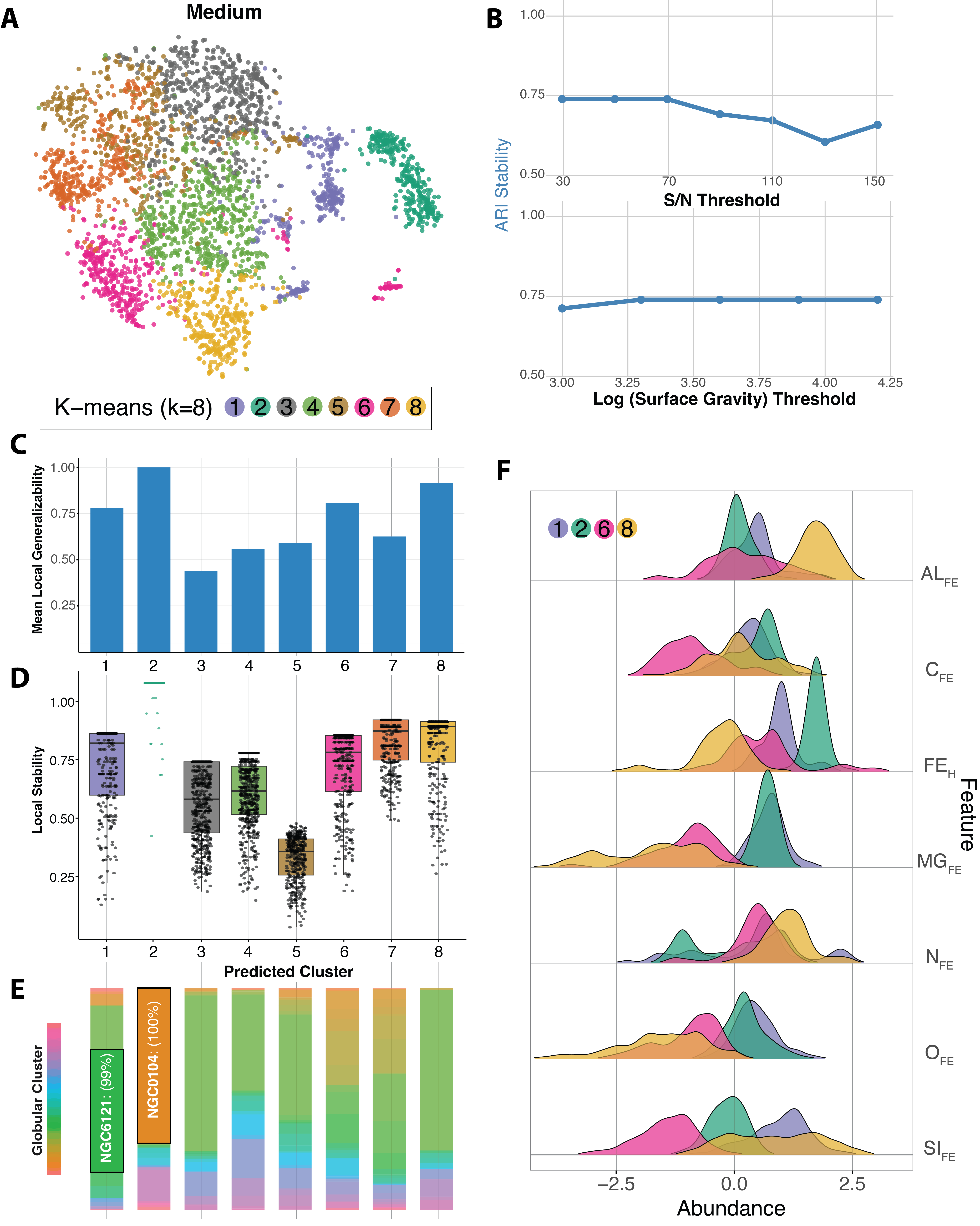}
    \caption{Final clustering of stars from APOGEE survey, refit on the entire dataset. \textbf{A:} The eight K-means clusters fit on the entire medium, mean-imputed dataset. \textbf{B:} Sensitivity of final clustering results to different astronomical preprocessing choices. When varying the S/N cutoff (top) or threshold for the log of surface gravity (bottom) of stars, we repeat the clustering pipeline and evaluate ARI stability between the resulting clusters and our original eight K-means clusters from panel A. Changing these parameters has little to no effect on the final stellar groupings, demonstrating the robustness of our workflow and clustering pipeline to arbitrary filtering choices. We test further parameters in IS.6 and observe similar results.}
    \label{fig:fig4}
\end{figure}
\makeatletter
  \setlength{\OLDfptop}{\@fptop}
  \setlength{\@fptop}{0pt}
\makeatother
\begin{figure}[p]\ContinuedFloat
  \caption[]{(Previous page.) \textbf{C:} Local generalizability of each of the eight K-means clusters, defined as the precision per cluster when training a classifier to predict clustering labels. Cluster 2 has local generalizability = 1.00 (see IS.6) across the entire pipeline, with cluster 3 having the lowest generalizability. \textbf{D:} Distribution of local stability scores for each cluster across imputation methods and datasets. Cluster 2 again has the highest local stability, whereas cluster 5 has the lowest. From C and D, we conclude that clusters 1, 2, 6 and 8 are the most generalizable and stable clusters. \textbf{E:} Globular cluster (GC) composition of each of the eight clusters. NGC6121 is nearly entirely contained within cluster 1 (98.8\%), and NGC0104 is entirely contained within cluster 2, a potential source of the strong stability rates for these clusters. \textbf{F:} From panels C, D, we conclude clusters 1, 2, 6, and 8 are the most stable and generalizable, and hence suitable for scientific interpretation and communication. Here, we interpret seven key chemical abundance distributions across these four clusters. Cluster 1 is iron- and silicon-high, whereas cluster 2, seen in the separated right branch in the tSNE embedding (A), is an iron-rich cluster likely made up of younger stellar bodies. Cluster 6 is carbon- and silicon-low, and cluster 8 is driven by aluminum enrichment and low levels of iron.}
\end{figure}
\makeatletter
  \setlength{\@fptop}{\OLDfptop}
\makeatother

\clearpage
\bibliographystyle{abbrv}
\bibliography{references}


\end{document}